\documentclass[9pt,twocolumn,twoside]{osajnl}
\usepackage{graphicx}
\usepackage{multirow}
\usepackage{indentfirst}
\usepackage{amsmath}
\usepackage{graphicx}
\usepackage{subfigure}

\journal{ol} 

\setboolean{shortarticle}{true}

\title{Towards Good Practices on Building Effective \\ CNN Baseline Model for Person Re-identification}

\author[1]{Fu Xiong}
\author[1,*]{Yang Xiao}
\author[1]{Zhiguo Cao}
\author[1]{Kaicheng Gong}
\author[2]{Zhiwen Fang}
\author[3]{Joey Tianyi Zhou}

\affil[1]{ National Key Laboratory of Science and Technology on Multi-Spectral Information Processing, School of Automation, Huazhong University of Science and Technology}
\affil[2]{ School of Energy and Mechanical-electronic Engineering, Hunan University of Humanities, Science and Technology, Loudi, China}
\affil[3]{ Institute of High Performance Computing, A*STAR, Singapore}

\affil[*]{Corresponding author:  Yang Xiao@hust.edu.cn}

\begin{abstract}
Person re-identification is indeed a challenging visual recognition task due to the critical issues of human pose variation, human body occlusion, camera view variation, etc. To address this, most of the state-of-the-art approaches are proposed based on deep convolutional neural network (CNN), being leveraged by its strong feature learning power and classification boundary fitting capacity. Although the vital role towards person re-identification, how to build effective CNN baseline model has not been well studied yet. To answer this open question, we propose 3 good practices in this paper from the perspectives of adjusting CNN architecture and training procedure. In particular, they are adding batch normalization after the global pooling layer, executing identity categorization directly using only one fully-connected, and using Adam as optimizer. The extensive experiments on 3 widely-used benchmark datasets demonstrate that, our propositions essentially facilitate the CNN baseline model to achieve the state-of-the-art performance without any other high-level domain knowledge or low-level technical trick.
\end{abstract}

\setboolean{displaycopyright}{true}

\begin{document}

\maketitle

\section{Introduction}
As a fine-grained visual recognition problem, person re-identification is of wide-range application scenarios, such as public security system, content-based image retrieval, etc. Under the practical application conditions, the variation on human pose and camera view, and human body occlusion impose great challenges to this task. To address this, numerous of efforts have been paid from the different theoretical perspectives~{\cite{Liao2014Person}}, such as seeking discriminative visual representation and defining applicable distance metric. The most recently, the introduction of deep convolutional neural network (CNN)~{\cite{Zheng2016Person}} leverages the performance significantly. CNN's main superiority is that, it can optimize the procedures of visual feature extraction, metric learning and classification jointly in end-to-end learning manner.

At the current stage, most of the state-of-the-art person re-identification approaches~{\cite{Zheng2016Person,Chang2018Multi,Si2018Dual}} are proposed based on CNN. Generally, fine-tuning the CNN model pre-trained on ImageNet~{\cite{deng2009imagenet}} under the supervision of softmax loss usually serves as the baseline paradigm. However, due to the variational implementation and training details the reported rank-1 results of CNN baseline model are often of high divergence (e.g., over 10\%) in different publications~{\cite{Chang2018Multi,Si2018Dual}}, even with the same CNN architecture. This phenomenon actually imposes negative impacts to the research community. First, it essentially leads to unfair comparison between the different approaches. Secondly, this leads to the fact that it is hard to judge the real capacity of the existing CNN-based person re-identification methods. Unfortunately, this problem has not been well studied yet. Thus, \emph{we argue that study on building effective CNN baseline model is urgently required} to benefit person re-identification research field to large extent, both from the theoretical and application perspectives. .

\begin{figure*}
\centering
\includegraphics[width=17cm,height=6cm]{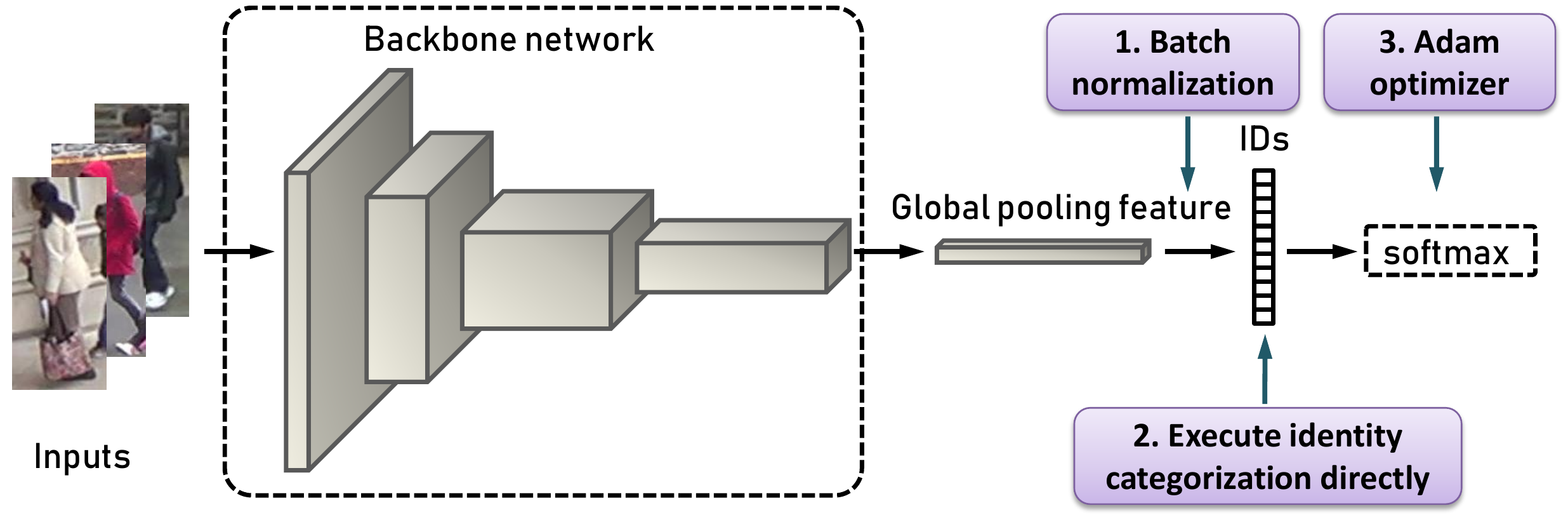}
\caption{The main technical pipeline our proposed CNN-based person re-identification approach.}
\label{fig:pipeline}
\end{figure*}

In this paper, we propose 3 good practices for building effective CNN baseline model towards person re-identification. Our main propositions include:

$\bullet$ Add a batch normalization layer after the global pooling layer to prevent overfitting;

$\bullet$ Directly use the batch-normalized global pooling feature to execute identity classification, using only one fully-connected layer;

$\bullet$ Employ Adam~{\cite{Adam}} as the optimizer for CNN training.

The experiments on 3 challenging datasets (i.e., Market-1501~{\cite{Zheng2016Scalable}}, DukeMTMC-reID~{\cite{Zheng2017Unlabeled}}, and CUHK03~{\cite{Zhong2017Re}}) demonstrate that, being facilitated with our propositions the commonly used CNN baseline models (e.g., Resnet50~{\cite{He2016Deep}}, Resnext50~{\cite{Xie2016Aggregated}}, and Densenet121~{\cite{Huang2016Densely}}) can easily achieve the state-of-the-art performance, only using softmax loss without any other high-level domain knowledge or low-level technical trick.

Actually, our proposition is a simple but effective way to achieve the state-of-the-art performance, and easy to reproduce. By shedding the light to building effective CNN baseline model, we believe that our contribution can essentially promote person re-identification research field. The source code and supporting materials of our proposition will be published online upon acceptance.

\section{Related works}
\label{sec:Related}
In this section we mainly focus on deep learning baseline for person re-identification, and existing Approaches for preventing over fitting. ~{\cite{Zheng2016Person}} is recommend if the readers are interested in an overall review of re-ID.
\subsection{Baseline for deep person re-identification }
Traditionally, hand-crafted feature designed based on color histogram prevails in person re-ID~\cite{Jurie2012PCCA,Liao2014Person,Gray2008Viewpoint}, due to the observation that color of clothes has good discriminability for distinguishing persons. Recent researches on person re-ID mostly focus on building deep convolutional neural network in the end-to-end learning manner. ~{\cite{Zheng2016Person}} takes advantage of the deep convolutional models pre-trained on ImageNet\cite{deng2009imagenet} and fine-tunes it on person re-identification datasets using softmax loss. The feature from the last pooling layer is used as the image representation.  The learned representation achieves great performance boost against traditional hand-crafted feature. Due to the success of ~{\cite{Zheng2016Person}}, most current deep learning based methods also adopt pre-trained models as backbone network and have been searching other technical means to further leverage the performance of re-ID system. Hence, for most current approaches ~{\cite{SVDnet,DaF,Chang2018Multi}} feature learned using only softmax loss usually serves as a baseline for comparison.

Different network architectures have been explored in person re-identification filed. Among existing person re-identification approaches, Resnet50~{\cite{He2016Deep}} is the most commonly used backbone network~{\cite{SVDnet,DaF}}. Besides, GoogLeNet~{\cite{googlenet}}, inception networks~{\cite{icpv3}} and densenet~{\cite{Huang2016Densely} have also been chosen as backbone network by some researchers. Taking advantage of the pre-trained CNN models, by further employing metric learning methods~\cite{Hermans2017In}, using part-based CNN representation\cite{DPA} or carefully designing attention mechanism\cite{Li2018Harmonious}, the performance for person re-identification can be further improved.

\subsection{Approaches for preventing overfitting}
To prevent overfitting for deep CNN models when trained on relatively small datasets, many approaches have been proposed. In particular, random cropping~\cite{randcrop}, random flipping~\cite{randflip} and random erase operation~\cite{Randomerase} are commonly used data augmentation methods in training deep CNN model. Besides, regularization methods like weight decay is also a well-known approach for prevent overfitting. Recently, batch normalization and dropout are two widely used tricks for training CNN, and have show benefits for preventing overfitting. Dropout randomly discards the output of each hidden neuron with a probability during the training process. Batch normalization aims at reducing internal co-variate shift by normalizing the output of each hidden neuron using mini-batch mean and variance. Since person re-identification dataset are relatively small (e.g Market1501 containing only 12,936 images for training), effective means for preventing overfitting is necessary for building high-accuracy person re-identification model.

\section{Good practices on building effective CNN baseline model}
In this section, we will illustrate 3 key practices towards building effective CNN baseline model for person re-identification. The main technical pipeline of out approach and our proposed 3 good practices are shown in Fig.~\ref{fig:pipeline}. First, the inputs are fed into the backbone CNN network (e.g., Resnet50). Then, the ``global pooling feature" yielded by the last pooling layer is fed into the batch normalization layer to generate the final subject representation. With the batch-normalized feature, we directly execute person identity categorization using only one fully-connected layer. That is, the procedure of dimension reduction using multiple fully-connected layers~{\cite{VGG,SVDnet}} is removed. At last, Adam is employed as the optimizer to train CNN model. Intuitively, our proposition is simple but can be easily applied to various kinds of CNN architectures. It is worthy noting that, only softmax loss is used as supervision signal in our approach. 

Although the 3 practices are proposed mainly according to our experience and experiments, we guess the superiority of these 3 practices lies on preventing overfitting and maintaining the discriminative power of the model pre-trained on ImageNet. ImageNet dataset contains over 14 million images, which has provided plenty of visual concepts. Recently, models pre-trained on ImageNet have demonstrated good transfer ability in various computer vision tasks~\cite{FCN,YOLOV2}. Hence, we think a carefully designed fine-tuning procedure can also benefit person re-ID by preserving the discriminative power of pre-trained model and preventing overfitting. Next, we will illustrate the proposed 3 good practices in details.

\subsection{Execute batch normalization after global feature pooling} \label{sec:theo_bn}
To well fine-tune CNN model pre-trained on ImageNet towards the relative small-scale datasets (e.g., person re-identification dataset), one critical issue is to alleviate overfitting problem during training. Currently, dropout~{\cite{Srivastava2014Dropout}} and batch normalization~{\cite{Ioffe2015Batch}} are two widely used technologies to address this. Dropout randomly discards the output of the hidden neuron with a probability during the training process. Batch normalization aims at reducing internal covariate shift by normalizing the output of each hidden neuron using mini-batch mean and variance.

For the specific person re-identification task, some works~\cite{Zheng2016A} adopt dropout after the global pooling layer in order to prevent overfitting. In our implementation, we empirically choose to execute batch normalization after the global feature pooling layer instead of dropout, as shown in Fig.~\ref{fig:pipeline}.  In the testing phase, the feature after the batch-normalization layer is chosen as the person image representation. Compared with dropout, which needs to randomly discards some hidden neuron, batch normalization can provide more steady gradient. This can alleviate the unnecessary disturbance to the pre-trained model.

In our experiments we find that, batch normalization generally leads to faster convergence speed and better performance. The experiments conducted in Sec.~\ref{sec:experi} will verify the effectiveness of our proposition.
\subsection{Conduct identity categorization directly using only one fully-connected layer}

Within some well-established person re-identification CNN models~{\cite{SVDnet}}, 2 fully-connected layers are usually set after the global feature pooling layer. The first fully-connected layer plays the role of ``bottleneck" to conduct feature dimension reduction. And, the second fully-connected layer executes person identity categorization. However, according to our experience the introduction of ``bottleneck" fully-connected layer often decreases the final performance essentially. Thus, we choose to remove the ``bottleneck" layer. With the batch normalized global pooling feature yielded in Sec.~\ref{sec:theo_bn}, we propose to conduct person identity categorization directly using only one fully-connected layer.

It is worth noting that, our proposition actually helps to compress the employed CNN model. By discarding the  ``bottleneck"  layer, the gradient from the softmax loss can be directly pass backward to the convolutional layer. This also essentially benefits to alleviate overfitting problem to ensure the test performance.

\subsection{Optimize CNN model using Adam}
The last key practice is to train CNN model using Adam as the optimizer, under the supervision of cross entropy loss. Adam is the recently proposed first-order gradient-based optimization method for stochastic objective function. It executes based on the adaptive estimates of lower-order moments. In particular, Adam update the parameters according to Eq.~\ref{con:ADAM}.
\begin{equation}
\label{con:ADAM}
\left\{ \begin{array}{l}
t \leftarrow t + 1\\
{g_t} \leftarrow {\nabla _\theta }{f_t}({\theta _{t - 1}})\\
{m_t} \leftarrow {\beta _1} \cdot {m_{t - 1}} + (1 - {\beta _1}) \cdot {g_t}\\
{v_t} \leftarrow {\beta _2} \cdot {v_{t - 1}} + (1 - {\beta _2}) \cdot {g_t}^2\\
{{\hat m}_t} \leftarrow {m_t}/(1 - {\beta _1}^t)\\
{{\hat v}_t} \leftarrow {v_t}/(1 - {\beta _2}^t)\\
{\theta _t} \leftarrow {\theta _{t - 1}} - \alpha  \cdot {{\hat m}_t}/(\sqrt {{{\hat v}_t}}  + \varepsilon ),
\end{array} \right.
\end{equation}
where t denotes the timestep. $f$ means the objective function. $\theta$ represents the learnable parameters. $\alpha$ denotes the learning rate. $\beta _1$, $\beta _2$ and $\varepsilon$ are hyper-parameters, whose defualt setting is $\beta _1 = 0.9$, $\beta _2 = 0.999$ and $\varepsilon=1e-8$.

Currently, most of the state-of-the-art person re-identification methods~{\cite{Si2018Dual,SVDnet}} choose stochastic gradient descent (SGD) to train CNN model. Compared with SGD, Adam adaptive estimates lower-order moments, which smooths the variation of the gradients. This can also alleviate the unnecessary disturbance to the pre-trained model as well as prevent over-fitting. The experiments conducted in Sec.~\ref{sec:experi} will verify the effectiveness of our proposition.
\section{Experiments} \label{sec:experi}
\subsection{Experimental setup}
\textbf{Datasets:} To evaluate the effectiveness of our propositions, we conduct experiments on three widely used person re-identification datasets, including Market-1501~{\cite{Zheng2016Scalable}}, DukeMTMC-reID~{\cite{Zheng2017Unlabeled}} and CUHK03~{\cite{Zhong2017Re}}. The Market-1501 dataset consists of image samples from 6 cameras with different resolutions. It contains 1,501 identities and 32,668 bounding boxes. The training set contains 12,936 bounding boxes of 751 identities. The remaining 750 identities are used for test. DukeMTMC-reID dataset contain 1,812 identities with 8 cameras. There are 1,404 identities that appear in more than two cameras, and 408 identities (i.e., distractor ID) that appear in only one camera. 702 identities are selected for training, and the remaining 702 identities as used for test. CUHK03 dataset consists of 14,097 image samples captured by 6 cameras from 1,467 persons. Two kinds of annotations are provided in this dataset: manually labeled pedestrian
bounding boxes and DPM-based \cite{DPM} bounding boxes. On CUHK03 dataset, we use the test protocol in \cite{Zhong2017Re}. That is, image samples from 767 identities are selected for training. The other 700 identities are employed for test. Fig.~\ref{fig:dataset} shows the image samples from these 3 datasets. It can be seen that, the issues of human pose variation, human body occlusion and camera view variation impose great challenge to this task.

\textbf{Evaluation:} In testing phase, Euclidean distance of $L_2$ normalized feautres is used as similarity metric. Cumulated Matching Characteristics (CMC) curve is employed to evaluate the performance of person re-identification methods. The CMC curve shows the probability that a query identity appears in different-sized candidate lists. Because of the space limitation , we only report the cumulated matching accuracy at selected ranks rather than plotting the actual curves. Mean average precision is also used as a performance measure. Mean average precision indicates the area under the Precision-Recapll curve.

\textbf{Implementation details:} Our CNN-based person re-identification approach is implemented using PyTorch framework. The CNN models used by us are pre-trained on ImageNet. The input image is resized to 256$\times$128 and padded to 276$\times$148 with zero. Input images are then randomly left-right flipped and cropped to 256$\times$128 for data augmentation. Left-right image flipping is also used in testing phase. Training iterations are set as 60 epochs for all datasets. The Adam \cite{Adam} optimiser is used with a mini-batch size of 32. Following \cite{Chang2018Multi}, the initial learning rate is set as 0.00035 and  weight decay is set as 5e-4 for all datasets. Adam parameters are set as $\beta_1$ = 0.9 and $\beta_2$ = 0.999. The decay factor for the learning rate is set as 0.1 at every 20 epochs.
\begin{figure}
\centering
\includegraphics[width=9cm,height=7cm]{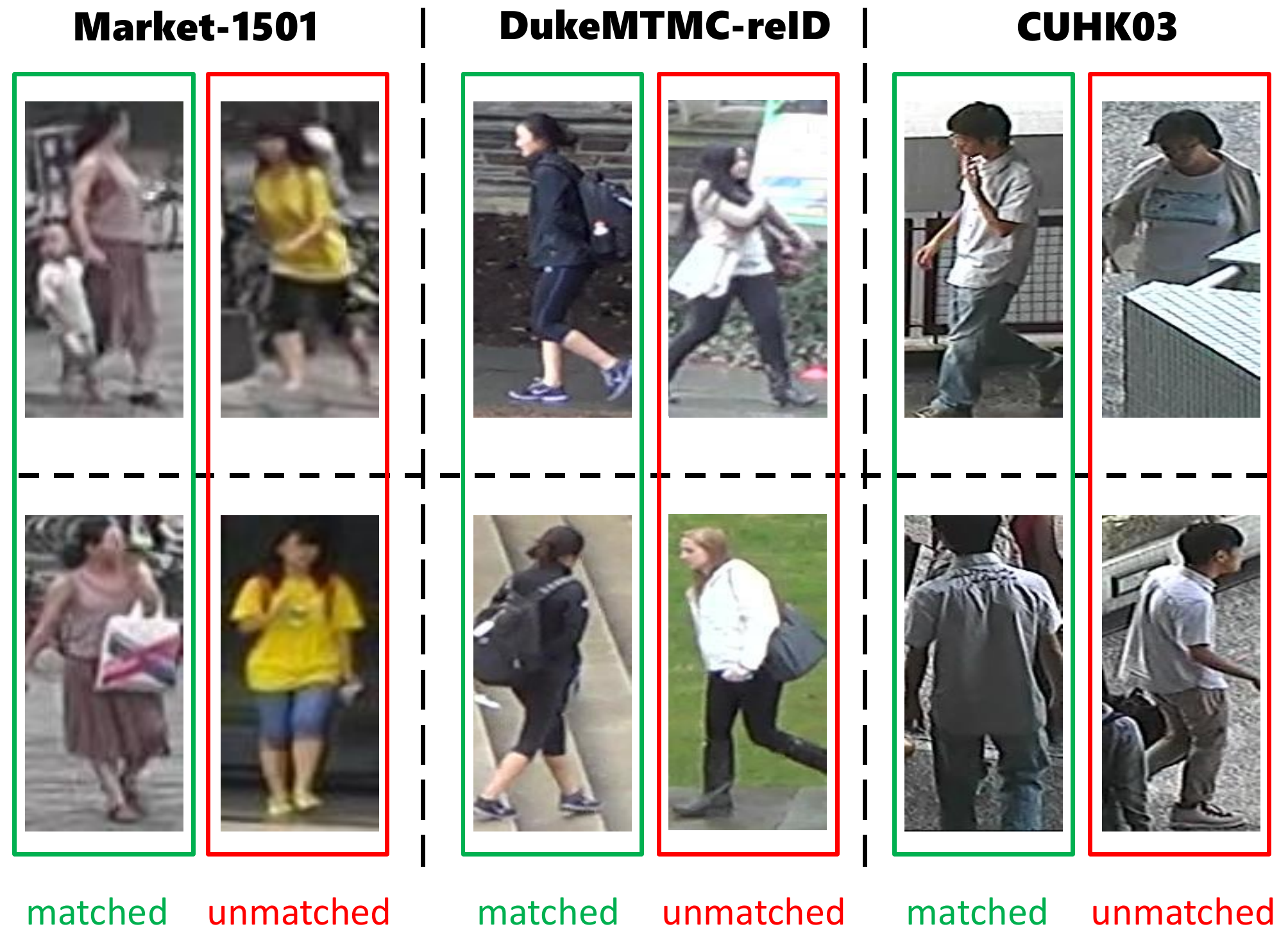}
\caption{Matched and unmatched sample pairs from Market-1501, DukeMTMC-reID and CUHK03 datasets. Matched pairs means the two images belong to the same identity. Unmatched pairs means the two images come from different identities.}
\label{fig:dataset}
\end{figure}

\subsection{Results on Market-1501 dataset}
Performance comparison between our method and the the other state-of-the-art approaches on Market-1501 dataset is shown in Table~\ref{table:Market}. Specifically, R50 represents Resnet50 network~{\cite{He2016Deep}}, R39 represents Resnet39 network~{\cite{He2016Deep}}, Rx50 represents Resnext50 network~{\cite{Xie2016Aggregated}}, Dse121 represents Densenet121 network~{\cite{Huang2016Densely}}, GN represents GoogLeNet network~{\cite{googlenet}}, and Icp-V3 represents Inception-V3 network~{\cite{icpv3}}. We do not report the CNN architecture if the corresponding method dose not use pre-trained model. To verify the effectiveness and generality of our proposition, it is applied to multiple CNN architectures (i.e., Resnet50, Resnext50 and Densenet121) widely used for person re-identification task. It can be observed that, being facilitated by the proposed good practices Resnet50, Resnext50 and Densenet121 outperform all the state-of-the-art approaches in all the test cases.

It is worthy noting that, DaRe\cite{Dare}, CamStyle+RE\cite{Camerastyle,RE}, MLFN\cite{Chang2018Multi}, HA-CNN\cite{Li2018Harmonious} and DuATM\cite{Si2018Dual} are the most recently published approaches at CVPR 2018. In particular, DaRe fuses the feature from the multiple layers of Resnet50. CamStyle+RE takes advantage of cycle-gan~{\cite{cyclegan}} to generate samples of different camera styles, and then uses them to fine-tuning CNN in the idea of data augment. MLFN, HA-CNN and DuATM design extra attention mechanism to acquire discriminative feature. However, we train CNN model only use softmax loss. No metric learning methods, no global local fusion and no attention mechanism are applied.

        \begin{table}
        \footnotesize
        \caption{Performance comparison on Market-1501 dataset.}
        \label{table:Market}
        \centering
        \begin{tabular}{|c|c|c|c|c|c|}
        \hline
        \multirow{2}*{Method} &\multicolumn{2}{|c|}{Single-query} &\multicolumn{2}{|c|}{Multi-query} &\multirow{2}*{Backbone}\\
        \cline{2-5}
        ~  &rank-1  &mAP &rank-1 &mAP &~\\
        \hline
        LOMO+XQDA~\cite{Liao2014Person} &43.8 &22.2 &54.1 &28.4&--\\
        IDE~\cite{Zheng2016Person} &73.9 &47.8 &-- &--&R50\\
        Spindle~\cite{Spindle} &76.9 &-- &-- &--&--\\
        OIM~\cite{OIM} &82.1 &-- &-- &--&R50\\
        Re-rank~\cite{Zhong2017Re} &77.1 &63.6 &-- &--&R50\\
        DPA~\cite{DPA} &81.0 &63.4 &-- &--&GN\\
        SVDNet~\cite{SVDnet} &82.3 &62.1 &-- &--&R50\\
        IDE+DaF~\cite{DaF} &82.3 &72.4 &-- &--&R50\\
        ACRN~\cite{ACRN} &83.6 &62.6 &-- &--&GN\\
        Context~\cite{context} &80.3 &57.5 &86.8 &66.7&--\\
        JLML~\cite{Li2017Person} &83.9 &64.4 &89.7 &74.5&R39\\
        SSM~\cite{SSM} &82.2 &68.8 &88.2 &76.2&R50\\
        DaRe~\cite{Dare} &86.4 &69.3&--&--&R50 \\
        DPFL~\cite{DPFL} &88.6 &72.6 &92.2 &80.4&Icp-V3\\
        CamStyle+RE~\cite{Camerastyle,RE} &89.5 &71.6&--&--& R50\\
        MLFN~\cite{Chang2018Multi} &90.0 &74.3 &92.3 &82.4&Rx50\\
        HA-CNN~\cite{Li2018Harmonious} &91.2 &75.7 &93.8 &82.8&--\\
        DuATM~\cite{Si2018Dual} &91.4 &76.6 &--&--&Dse121\\
        \hline
        Resnet50 (ours) &\textbf{91.7} &\textbf{78.8}&\textbf{94.5}&\textbf{85.3}&R50\\
        Resnext50 (ours) &\textbf{92.0} &\textbf{78.9}&\textbf{94.1}&\textbf{85.4}&Rx50\\
        Densenet121 (ours) &\textbf{92.5} &\textbf{79.8} &\textbf{95.0}&\textbf{86.2}&Dse121\\
        \hline
        \end{tabular}
        \end{table}

        \begin{table}
        \footnotesize
        \caption{Performance comparison on DukeMTMC-reID dataset.}
        \label{table:Duke}
        \centering
        \begin{tabular}{|c|c|c|c|}
        \hline
        Methods &Rank-1 &mAP&Backbone\\
        \hline
        BoW+kissme~\cite{Zheng2016Scalable}  &25.1  &12.2&--\\
        LOMO+XQDA~\cite{Liao2014Person}  &30.8 & 17.0&--\\
        IDE~\cite{Zheng2016Person} &65.2 &45.0&R50\\
        ACRN~\cite{ACRN} &72.6 &52.0&GN\\
        SVDNet~\cite{SVDnet} &76.7 &56.8&R50\\
        AACN~\cite{AACN} &76.8 &59.3&GN\\
        CamStyle+RE~\cite{Camerastyle,RE} &78.3 &57.6& R50\\
        DPFL~\cite{DPFL} &79.2 &60.6&Icp-V3\\
        PSE~\cite{Sarfraz2017A} &79.8 &62.0&R50 \\
        ATWL~\cite{ATWL} &79.8 &63.4& R50\\
        DaRe~\cite{Dare} &75.2 &57.4&R50\\
        HA-CNN~\cite{Li2018Harmonious} &80.5 &63.8&--\\
        MLFN~\cite{Chang2018Multi} &81.2 &62.8&Rx50\\
        DuATM~\cite{Si2018Dual} &81.8 &64.6&Dse121\\
        \hline
        Resnet50 (ours) &\textbf{83.4} &\textbf{68.8}&R50\\
        Resnext50 (ours) &\textbf{82.8} &\textbf{68.1}&Rx50\\
        Densenet121 (ours) &\textbf{83.5} &\textbf{68.5}&Dse121\\
        \hline
        \end{tabular}
        \end{table}

\subsection{Results on DukeMTMC-reID dataset}
Performance comparison between our method and the the other state-of-the-art approaches on DukeMTMC-reID dataset is shown in Table~\ref{table:Duke}. We can see that the proposed method still outperforms all the state-of-the-art approaches in all the test cases. And, the superiority of our method on mAP is notable. To the best of our knowledge, our approach is the simplest one able top achieve the state-of-the-art performance.

\subsection{Results on CUHK03 dataset}
Table~\ref{table:CUHK03} lists the performance comparison between our approach and the state-of-the-art ones on CUHK03 dataset. It can be observed that, the superiority of our proposition is remarkable. This demonstrates the effectiveness and generality of the 3 proposed practices.

        \begin{table}
        \footnotesize
        \caption{Performance Comparison on CUHK03 dataset.}
        \label{table:CUHK03}
        \centering
        \begin{tabular}{|c|c|c|c|c|c|}
        \hline
        \multirow{2}*{Method} &\multicolumn{2}{|c|}{labeled} &\multicolumn{2}{|c|}{detected} &\multirow{2}*{Backbone}\\
        \cline{2-5}
        ~  &rank-1  &mAP &rank-1 &mAP &~\\
        \hline
        BOW+XQDA~\cite{Zheng2016Scalable} &7.9 &7.3 &6.4 &6.4&--\\
        LOMO+XQDA~\cite{Liao2014Person} &14.8 &13.6 &12.8 &11.5&--\\
        IDE~\cite{Zheng2016Person} &22.2 &21.0 &21.3 &19.7&R50\\
        IDE+DaF~\cite{DaF} &27.5 &31.5 &26.4 &30.0&R50\\
        DPFL~\cite{DPFL} &43.0 &40.5 &40.7 &37.0&Icp-V3\\
        SVDNet~\cite{SVDnet} &40.9 &37.8 &41.5 &37.3&R50\\
        HA-CNN~\cite{Li2018Harmonious} &44.4 &41.0 &41.7 &38.6&--\\
        MLFN~\cite{Chang2018Multi} &54.7 &49.2 &52.8 &47.8&Rx50\\
        DaRe~\cite{Dare} &58.1 &53.7 &55.1 &51.3&R50\\
        \hline
        Resnet50(ours) &\textbf{62.1} &\textbf{58.1}&\textbf{56.5}&\textbf{52.2}&R50\\
        Resnext50(ours) &\textbf{60.1} &\textbf{56.2} &\textbf{59.1}&\textbf{54.4}&Rx50\\
        Densenet121(ours) &\textbf{63.5} &\textbf{59.0}&\textbf{57.8}&\textbf{52.8}&Dse121\\
        \hline
        \end{tabular}
        \end{table}

\subsection{Ablation study}
In this subsection, we will conduct extensive ablation study on the 3 proposed practices to verify their effectiveness respectively. In particular, Resnet50 is chosen as the backbone network. And, Market-1501 and DukeMTMC-reID are employed as the test datasets.
        \begin{table}
        \footnotesize
        \caption{Ablation study on practice 1.}
        \label{table:as1}
        \centering
        \begin{tabular}{|c|c|c|c|c|}
        \hline
        \multirow{2}*{Dataset} &\multicolumn{2}{|c|}{Market-1501} &\multicolumn{2}{|c|}{DukeMTMC-reID} \\
        \cline{2-5}
        ~  &rank-1  &mAP &rank-1 &mAP \\
        \hline
        w/o BN &78.6 &57.9 &70.0 &50.9\\
        Dropout &79.2 &59.9 &72.9 &55.4\\
        \hline
        Good practices &\textbf{91.7} &\textbf{78.8}&\textbf{83.4} &\textbf{68.8}\\
        \hline
        \end{tabular}
        \end{table}

\textbf{Ablation study on adding batch normalization layer:} To verify the effectiveness of practice 1, we conduct two kinds of experiments. First, we directly leave out the batch normalization layer (denoted as w/o BN in Table.~\ref{table:as1}) and see how the network performs. Secondly, instead of directly leaving out the batch normalization layer, we switch the batch normalization layer to dropout and set the dropout ratio as 0.5 (denoted as Dropout in Table.~\ref{table:as1}). The performance comparison is reported in Table.~\ref{table:as1}. It can be seen that whether using batch normalization after the global pooling layer affects the final performance significantly. The mAP will drop from 78.8\% to 57.9\% on Market-1501 dataset and from 68.8\% to 50.9\% on DukeMTMC-reID dataset, if we leave out the batch normalization layer. Although dropout can enhance the performance, it is remarkably inferior to batch normalization. It is worthy noting that, dropout only achieves mAP of 59.9\% on Market-1501 dataset, and 55.4\% on DukeMTMC-reID dataset.

\textbf{Ablation study on executing identity classification directly:} To verify the effectiveness of practice 2, we insert another fully-connected layer as ``bottleneck" between the global pooling layer and the batch normalization layer. The feature after the batch normalization layer is used as the visual representation (denoted as Bottleneck in Table Table.~\ref{table:as2}). The bottleneck dimension is set as 512. Whether using ``bottleneck" also largely affects the final performance. In our experiments, we find that ``bottleneck" layer indeed weakens the performance. mAP will drop from 78.8\% to 73.4\% on Market-1501 dataset, and drop from 68.8\% to 61.8\% on DukeMTMC-reID dataset, if ``bottleneck" layer is used.
        \begin{table}
        \footnotesize
        \caption{Ablation study on practice 2.}
        \label{table:as2}
        \centering
        \begin{tabular}{|c|c|c|c|c|}
        \hline
        \multirow{2}*{Dataset} &\multicolumn{2}{|c|}{Market-1501} &\multicolumn{2}{|c|}{DukeMTMC-reID} \\
        \cline{2-5}
        ~  &rank-1  &mAP &rank-1 &mAP \\
        \hline
        Bottleneck &89.4 &73.4 &78.6 &61.8\\
        \hline
        Good practices &\textbf{91.7} &\textbf{78.8}&\textbf{83.4} &\textbf{68.8}\\
        \hline
        \end{tabular}
        \end{table}

\textbf{Ablation study on using Adam:} To verify the effectiveness of practice 3, we replace Adam optimizer with SGD optimizer. And, the initial learning rate for SGD is set as 0.01 and the momentum is set as 0.9 (denoted as SGD in Table.~\ref{table:as3}). The learning rate decay is set as 0.1 every 20 epochs. The corresponding results are reported in Table.~\ref{table:as3} . As we can see that, Adam outperforms SGD consistently. The mAP will drop from 78.8\% to 72.4\% on Market-1501 dataset, and drop from 68.8\% to 64.5\% on DukeMTMC-reID dataset, if SGD is used as the optimizer for CNN model training.
        \begin{table}
        \footnotesize
        \caption{Ablation study on practice 3.}
        \label{table:as3}
        \centering
        \begin{tabular}{|c|c|c|c|c|}
        \hline
        \multirow{2}*{Dataset} &\multicolumn{2}{|c|}{Market-1501} &\multicolumn{2}{|c|}{DukeMTMC-reID} \\
        \cline{2-5}
        ~  &rank-1  &mAP &rank-1 &mAP \\
        \hline
        SGD &88.9 &72.4 &80.0 &64.5\\
        \hline
        Good practices &\textbf{91.7} &\textbf{78.8}&\textbf{83.4} &\textbf{68.8}\\
        \hline
        \end{tabular}
        \end{table}
        
\subsection{Failure cases}
In this section we will illustrate the failure cases under the rank-1 measurement. The experiment is conducted on DukeMTMC-reID dataset using Resnet50. We believe this experiment can help to figure out what is the major challenge of person re-ID for state-of-the-art person re-identification model. According to our observation, we categorize the failure cases into four categories, they are: (1) failure cases caused by occlusion; (2) failure cases caused by identities with similar-looking (3) failure cases caused by multi-person query images. (4) inexplicable failure cases. next we will show them respectively. To quantify the four cases, We have also counted the percentage of each case, which is shown in Table~\ref{table:failure cases}.
        \begin{table}
        \footnotesize
        \caption{Percentage of each category of the failure cases.}
        \label{table:failure cases}
        \centering
        \begin{tabular}{|c|c|c|c|c|}
        \hline
        Fail cases  &category 1  &category 2 &category 3 &category 4 \\
        \hline
        Percentage &19.6\% &34.3\% &37.8\% &8.3\%\\
        \hline
        \end{tabular}
        \end{table}

\textbf{Failure cases 1 -- caused by occlusion}: When the person in the image, especially the query image, is heavily occluded, the accuracy of person re-identification drops significantly. The failure cases of this category are illustrated in figure~\ref{fig:case1}. 
\begin{figure}
\centering
\includegraphics[width=9cm,height=5.4cm]{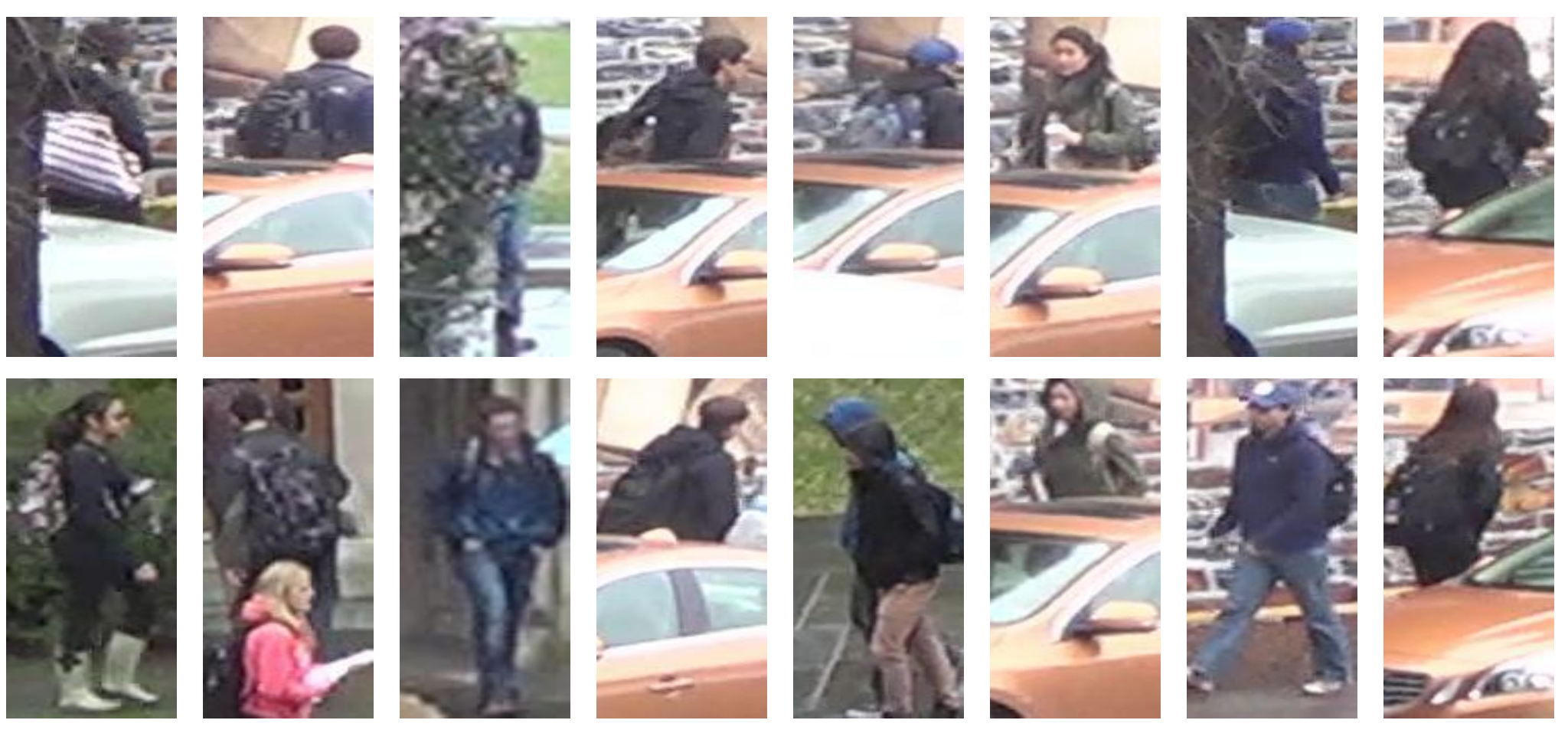}
\caption{Failure cases 1: failure cases caused by occlusion.}
\label{fig:case1}
\end{figure}

Occlusion affects the performance in two aspects. First, occlusion makes the description of the person not reliable, because the useful information only comes from the visible part of the body. Secondly, since the representation is obtained using global pooling, the feature of the occlusion objects may also be encoded into the representation. This leads to inaccurate description of the person. This category of failure cases account  for 19.6\% of all failure cases.

\textbf{Failure cases 2 -- caused by similar looking:}
How to identify persons with similar looking is an inherent difficulty in person re-identification. Actually, our model achieves rank-1 accuracy of 83.4\%. So the failure cases illustrated in fig~\ref{fig:case2} will tell us what kind of images pairs are still challenging for the state-of-the-art person re-identification model. This category of failure cases account for 34.3\% of all failure cases, which we think is the major challenge of person re-identification.
\begin{figure}
\centering
\includegraphics[width=9cm,height=5.4cm]{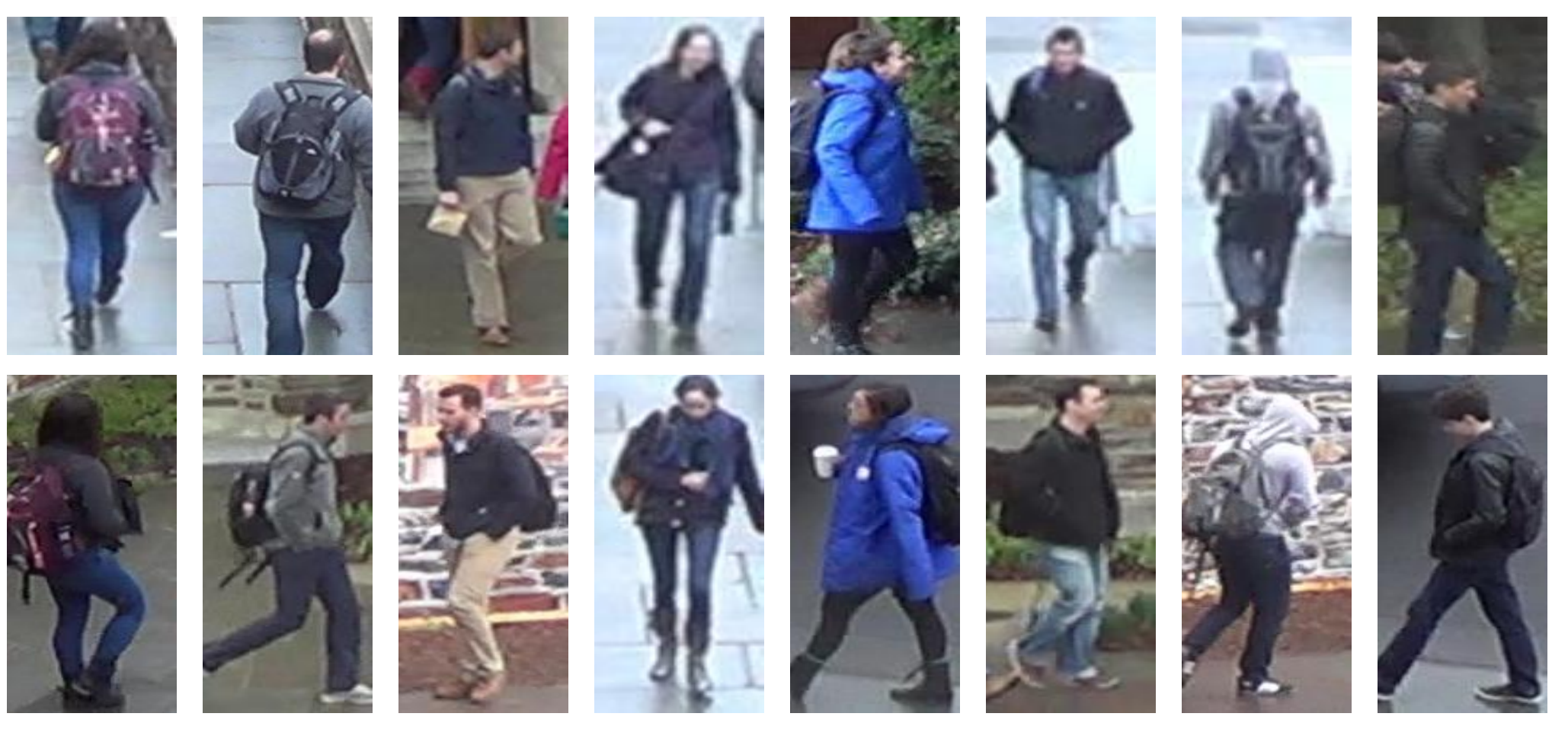}
\caption{Failure case 2: different person with similar looking.}
\label{fig:case2}
\end{figure}

\textbf{Failure cases 3 -- multiple persons in one query images:}
The third category of failure cases arise from multiple person in one query image, which is illustrated in Fig~\ref{fig:case3}. We think this kind of failure is tolerable though it account most for the failure cases, because in this case it is ambiguous that which person is the one for query. And in many failure cases, in spite that the searched person are not the person for query, it is indeed in the query image.

\textbf{Failure cases 4 -- inexplicable failure cases:}
The failure cases of this category are illustrated in Fig~\ref{fig:case4}. It is hard to name the cause of this category of failure cases. These failure cases are not hard to discriminate for human beings, but the CNN model fails to identify them. Hence, we call it inexplicable failure cases. Good news is that this category of failure cases account the least, only 8.3\%, for of all failure cases.

\section{Conclusions}
In this paper, we have proposed 3 good practices for building effective CNN baseline model towards person re-identification, including adding batch normalization layer after the global pooling layer, conducting identity categorization directly using only one fully-connected layer and using Adam as optimizer. By applying these 3 practices, we have achieved state-of-the-art performance on 3 challenging datasets. We believe that our proposition can serve as a good baseline to promote the research on person re-identification in future.

\section*{Acknowledgment}
This work is jointly supported by the National Key R\&D Program of China (No. 2018YFB1004600), National Natural Science Foundation of China (Grant No. 61502187, 61602193 and 61702182), the International Science \& Technology Cooperation Program of Hubei Province, China (Grant No. 2017AHB051), the HUST Interdisciplinary Innovation Team Foundation (Grant No. 2016JCTD120), Hunan Provincial Natural Science Foundation of China (Grant 2018JJ3254).

\begin{figure}
\centering
\includegraphics[width=9cm,height=5.4cm]{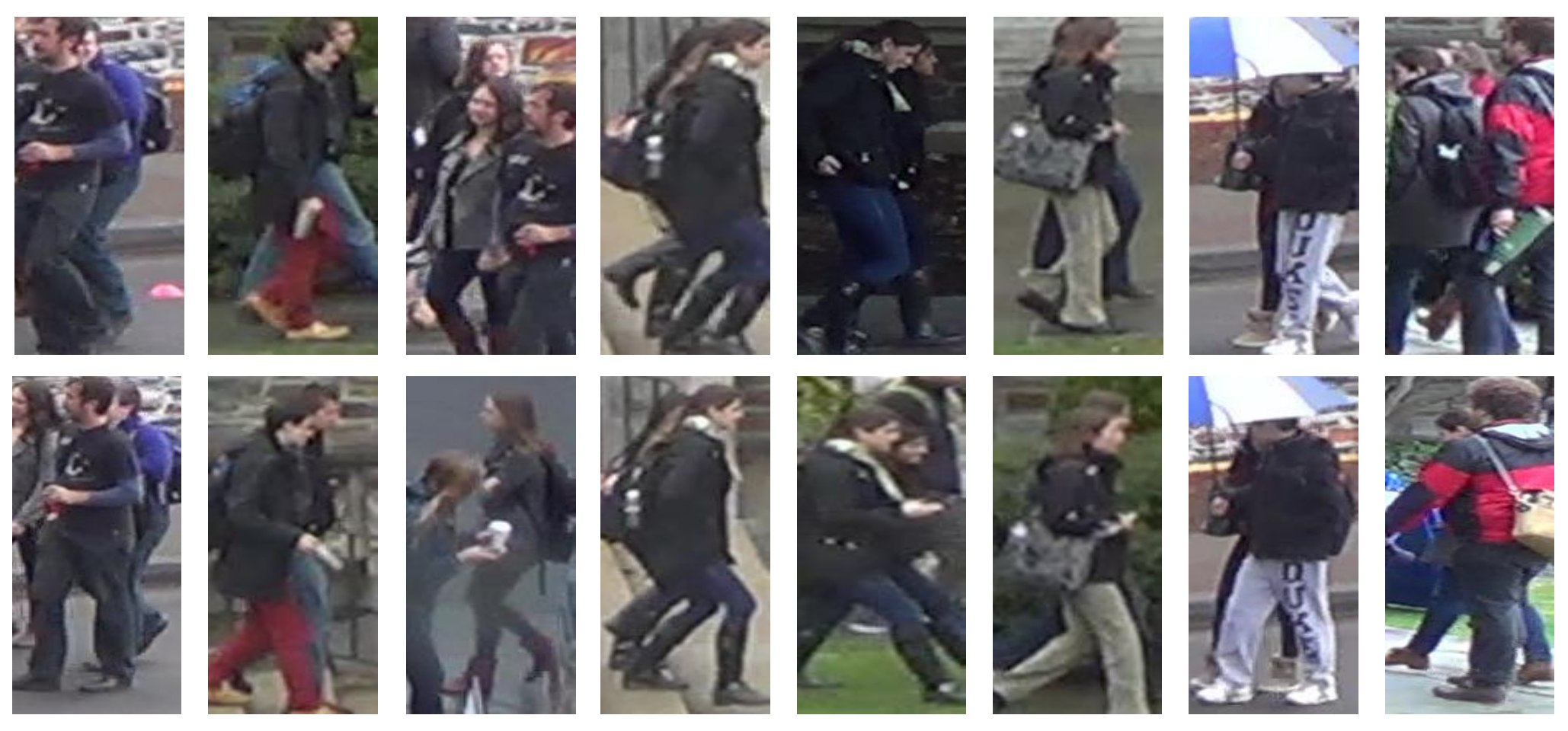}
\caption{Failure case 3: multiple persons in one query image.}
\label{fig:case3}
\end{figure}
\begin{figure}
\centering
\includegraphics[width=9cm,height=5.4cm]{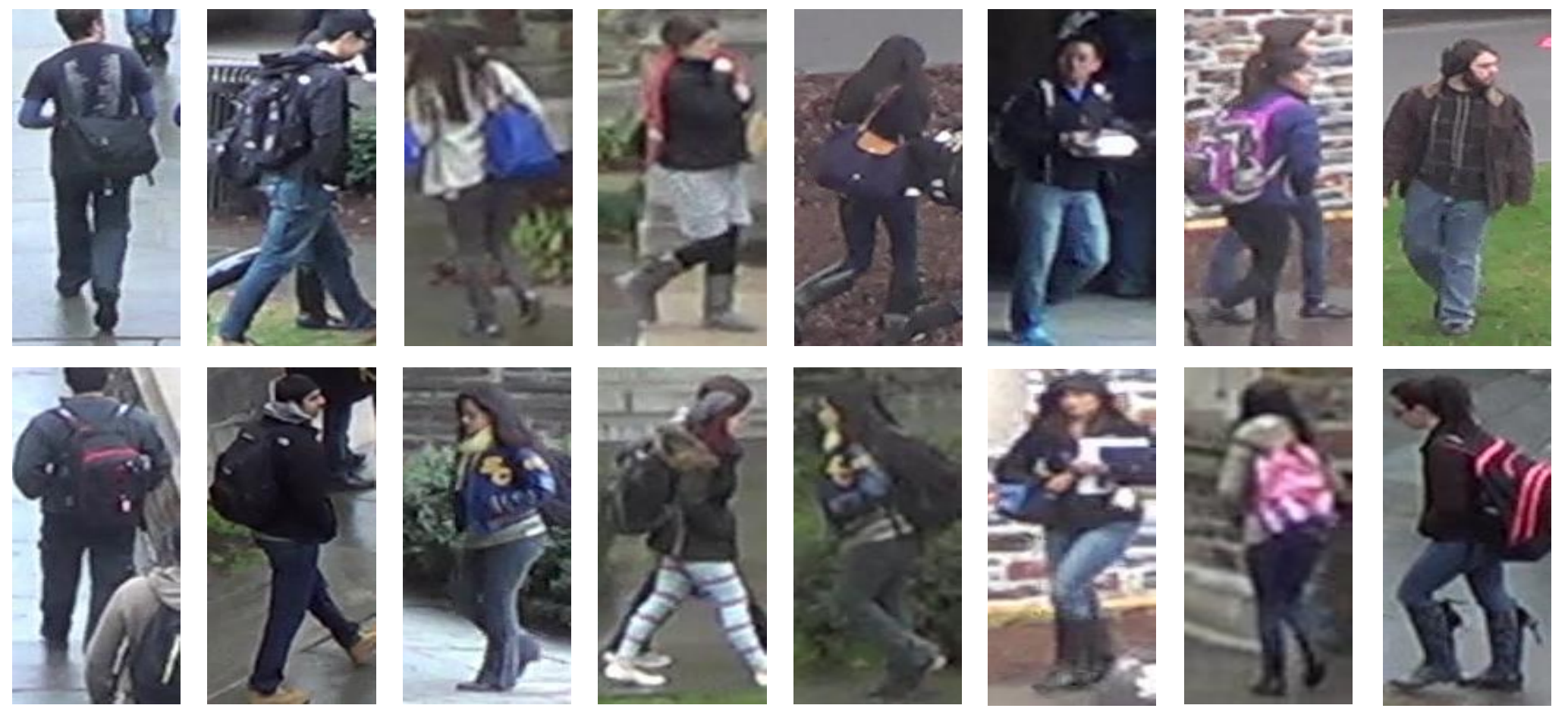}
\caption{Failure case 4: inexplicable failure cases.}
\label{fig:case4}
\end{figure}

\bibliography{sample}



\ifthenelse{\equal{\journalref}{aop}}{%
\section*{Author Biographies}
\begingroup
\setlength\intextsep{0pt}
\begin{minipage}[t][6.3cm][t]{1.0\textwidth} 
  \begin{wrapfigure}{L}{0.25\textwidth}
    \includegraphics[width=0.25\textwidth]{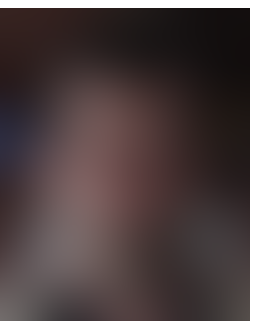}
  \end{wrapfigure}
  \noindent
  {\bfseries John Smith} received his BSc (Mathematics) in 2000 from The University of Maryland. His research interests include lasers and optics.
\end{minipage}
\begin{minipage}{1.0\textwidth}
  \begin{wrapfigure}{L}{0.25\textwidth}
    \includegraphics[width=0.25\textwidth]{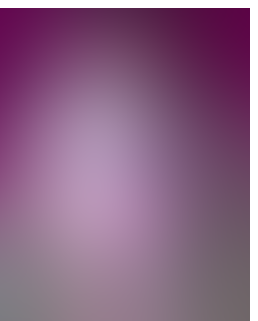}
  \end{wrapfigure}
  \noindent
  {\bfseries Alice Smith} also received her BSc (Mathematics) in 2000 from The University of Maryland. Her research interests also include lasers and optics.
\end{minipage}
\endgroup
}{}

\end{document}